\documentclass[sigconf]{acmart}
\usepackage{amsmath,amsfonts}
\usepackage{algorithmic}
\usepackage{graphicx}
\usepackage{textcomp}
\usepackage{xcolor}
\usepackage{multirow}

\usepackage{enumitem}

\usepackage{xspace}
\usepackage{bm}
\usepackage{subfigure}

\AtBeginDocument{%
  \providecommand\BibTeX{{%
    \normalfont B\kern-0.5em{\scshape i\kern-0.25em b}\kern-0.8em\TeX}}}

\copyrightyear{2023} 
\acmYear{2023} 
\setcopyright{acmlicensed}\acmConference[SIGIR '23]{Proceedings of the 46th International ACM SIGIR Conference on Research and Development in Information Retrieval}{July 23--27, 2023}{Taipei, Taiwan}
\acmBooktitle{Proceedings of the 46th International ACM SIGIR Conference on Research and Development in Information Retrieval (SIGIR '23), July 23--27, 2023, Taipei, Taiwan}
\acmPrice{15.00}
\acmDOI{10.1145/3539618.3592088}
\acmISBN{978-1-4503-9408-6/23/07}

\newcommand{\paratitle}[1]{\vspace{1.5ex}\noindent\textbf{#1}}
\newcommand{\ie}{\emph{i.e.,}\xspace}

\newcommand{\eg}{\emph{e.g.,}\xspace}

\newcommand{\wrt}{\emph{w.r.t.}\xspace}

\newcommand{\model}{MGDL}

\begin{document}
% \fancyhead{}

%%
%% The "title" command has an optional parameter,
%% allowing the author to define a "short title" to be used in page headers.
\title{Which Matters Most in Making Fund Investment Decisions? A Multi-granularity Graph Disentangled Learning Framework}

\author{Chunjing Gan}
\authornotemark[1]
\affiliation{
\institution{Ant Group}
\country{}
}
\email{cuibing.gcj@antgroup.com}

\author{Binbin Hu}
\authornotemark[1]
\affiliation{
\institution{Ant Group}
\country{}
}
\email{bin.hbb@antfin.com}
\thanks{* Equal contribution.}

\author{Bo Huang}
\affiliation{
\institution{Ant Group}
\country{}
}
\email{yunpo.hb@antgroup.com}

\author{Tianyu Zhao}
\affiliation{
\institution{Beijing University of Posts and Telecommunications}
\country{}
}
\email{tyzhao@bupt.edu.cn}

\author{Yingru Lin}
\affiliation{
\institution{Ant Group}
\country{}
}
\email{linyingru.lyr@alibaba-inc.com}

\author{Wenliang Zhong}
\affiliation{
\institution{Ant Group}
\country{}
}
\email{yice.zwl@antgroup.com}

\author{Zhiqiang Zhang}
\affiliation{
\institution{Ant Group}
\country{}
}
\email{lingyao.zzq@antfin.com}

\author{Jun Zhou}
\authornotemark[2]
\affiliation{
\institution{Ant Group}
\country{}
}
\email{jun.zhoujun@antfin.com}
\thanks{$^{\dagger}$ Corresponding author.}

\author{Chuan Shi}
\affiliation{
\institution{Beijing University of Posts and Telecommunications}
\country{}
}
\email{shichuan@bupt.edu.cn}

\renewcommand{\shortauthors}{Chunjing Gan et al.}

%%
%% By default, the full list of authors will be used in the page
%% headers. Often, this list is too long, and will overlap
%% other information printed in the page headers. This command allows
%% the author to define a more concise list
%% of authors' names for this purpose.
% \renewcommand{\shortauthors}{Trovato and Tobin, et al.}

%%
%% The abstract is a short summary of the work to be presented in the
%% article.
\begin{abstract}
In this paper, we highlight that both conformity and risk preference matter in making fund investment decisions beyond personal interest and seek to jointly characterize these aspects in a disentangled manner. Consequently, we develop 
a novel \underline{M}ulti-granularity \underline{G}raph \underline{D}isentangled
\underline{L}earning framework named {\model} to effectively perform intelligent matching of fund investment products. 
Benefiting from 
the well-established fund graph and the attention module, multi-granularity user representations are derived from historical behaviors to separately express personal interest, conformity and risk preference in a fine-grained way. To attain stronger disentangled representations with specific semantics, {\model} explicitly involve two self-supervised signals, \ie fund type based contrasts and fund popularity. 
Extensive experiments in offline and online environments verify the effectiveness of {\model}. 
\end{abstract}

\begin{CCSXML}
<ccs2012>
   <concept>
       <concept_id>10002951.10003317.10003347.10003350</concept_id>
       <concept_desc>Information systems~Recommender systems</concept_desc>
       <concept_significance>500</concept_significance>
       </concept>
 </ccs2012>
\end{CCSXML}

\ccsdesc[500]{Information systems~Recommender systems}

	\keywords{Graph Learning, Intelligent Matching, Disentangled Learning}

\maketitle

\section{Introduction}

The beginning of the economic era centered on ``personal finance'' encourages the flourishing of online investment platforms(\eg Wealthfront and Alipay). To help individual investors make fund investment decisions,  current financial platforms strive to provide intelligent
matching of fund products among a large number of choices, which can be 
naturally 
abstracted as a 
classical 
matching or recommendation problem~\cite{shi2018heterogeneous,zhang2019deep,hou2022core} with great interest-oriented efforts~\cite{shan2016deep,wang2021survey} based on sequential~\cite{kang2018self,sun2019bert4rec,fan2021lighter} and graph learning~\cite{hu2018leveraging,he2020lightgcn,fan2019graph,wang2019kgat,zou2022multi,fan2022graph,yang2022knowledge,liu2022unify} based modelling.
Despite considerable success in various traditional recommendation scenarios, \eg E-commerce, intelligent fund matching may be unlikely to benefit since personal interest may lose its leading role in the decision of financial products.

\begin{figure}[t]\centering
\includegraphics[width=7.5cm]{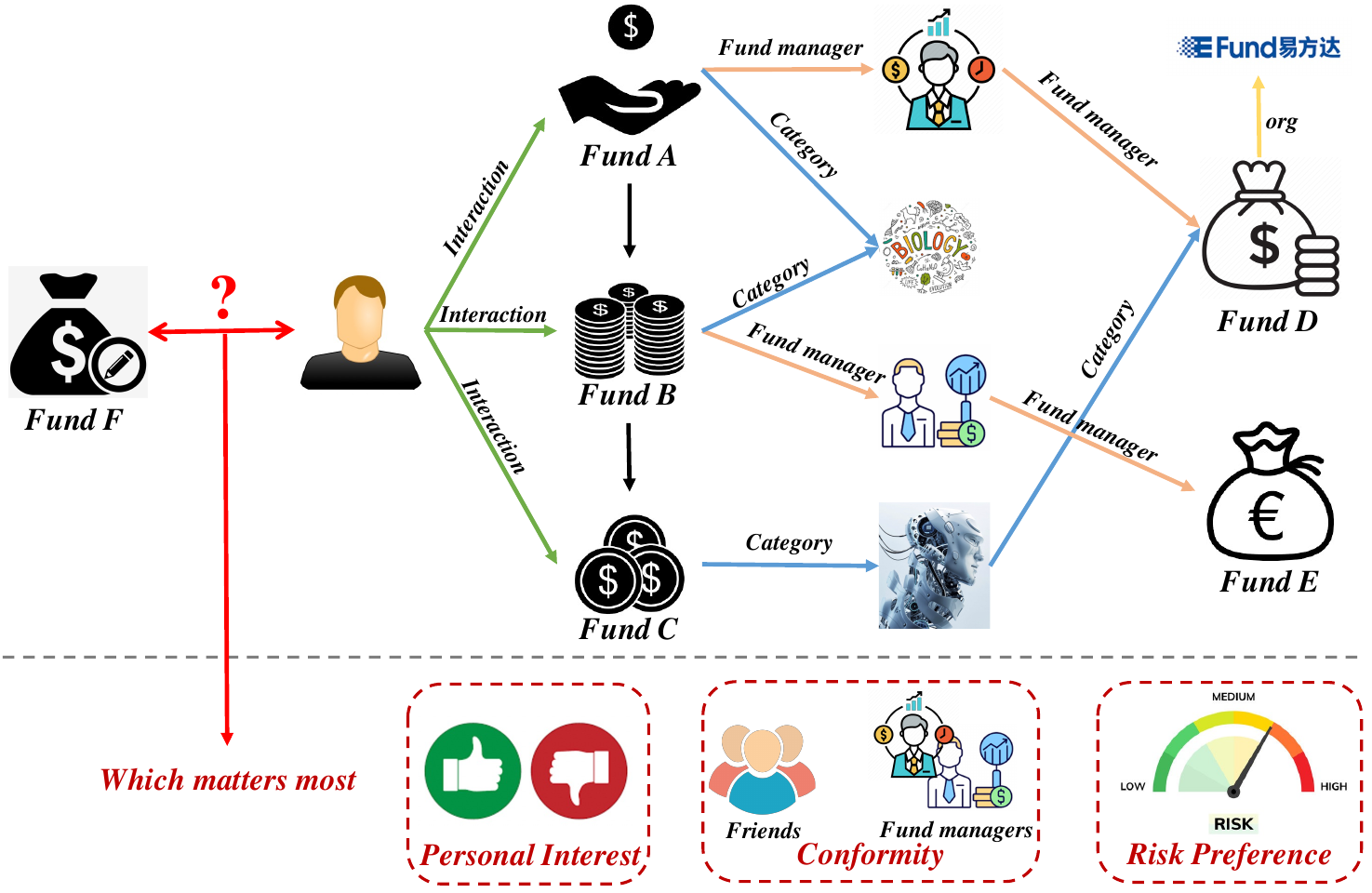}
\caption{A toy example of graph based intelligent fund matching in practical financial platforms.}
\label{fig:toy_example}
% \vspace{-1.5em}
\end{figure}

Comprehensive facts have shed 
light on the 
question ``Which matters most in making fund investment decisions beyond personal interest'', lying in the following two aspects related to the fairly unique financial scenarios (as shown in Fig.~\ref{fig:toy_example}):  
(1) \emph{Conformity} widely exists among individual investors.
In the current fund market, a wealth of investment products have sprung up. Unfortunately, most users' financial knowledge could 
not meet
their increasing investment needs, resulting in the common phenomenon that a large number of users buy fund products with the crowd. 
(2) \emph{Risk Preference} is of crucial importance for making investment decisions. 
Different fund products refer to different risk levels.
Therefore, users' risk preference derived from historical behavior, as a decisive signal, deserves more attention for discovering desired funds.

Intuitively, the idea of injecting both conformity and risk preference is impressive, while the solution is non-trivial, facing the following challenges. 
(\textbf{C1}): Users' investment decisions are attributed to multiple aspects, \ie personal interest, conformity and risk preference. Therefore, it is desired to develop a multi-granularity framework for disentanglement since a unified user representation is insufficient to capture such differences.
(\textbf{C2}): The interactivity between funds is powerful to capture users' disentangled representations, since fund products with similar categories or fund managers always show similar representations through interaction. Subsequently, high-order correlations between fund products are encouraged to be incorporated.
(\textbf{C3}): In the practical scenarios, only implicit feedback (\eg click) could be collected for guiding the overall learning procedure (\ie personal interest). Hence, it is hard to obtain external labeled data to distinguish the remaining aspects (\ie conformity and risk preference) with explicit supervision.

To tackle these challenges,  we propose \textbf{\model}, a  \underline{M}ulti-granularity \underline{G}raph \underline{D}isentangled
\underline{L}earning framework to help users discover the most proper fund products. 
To distinguish multiple aspects of user representations, we seek to build {\model} upon recently emerging disentangled procedure with historical behaviors, where multi-granularity representation could be obtained based on the attention mechanism in a fine-grained manner (\textbf{C1}).
By introducing the fund knowledge graph (Fig.~\ref{fig:toy_example}), we inject graph learning into sequential learning based on the well-designed fund graph, whose goal is to pull similar funds closer in the disentangled process while dynamic preference could be also summarized simultaneously (\textbf{C2}).
Aiming at alleviating the dependency on labeled data for learning multi-granularity user representations,
we creatively explore and explicitly exploit two parts of self-supervised signals:  fund type based contrasts
and fund popularity. (\textbf{C3}).
Multifaceted experiments show the superiority of {\model} across offline and online settings.

\section{The Proposed Approach}

\begin{figure}[t]
% \vspace{-1em}
\centering
\includegraphics[width=7.0cm]{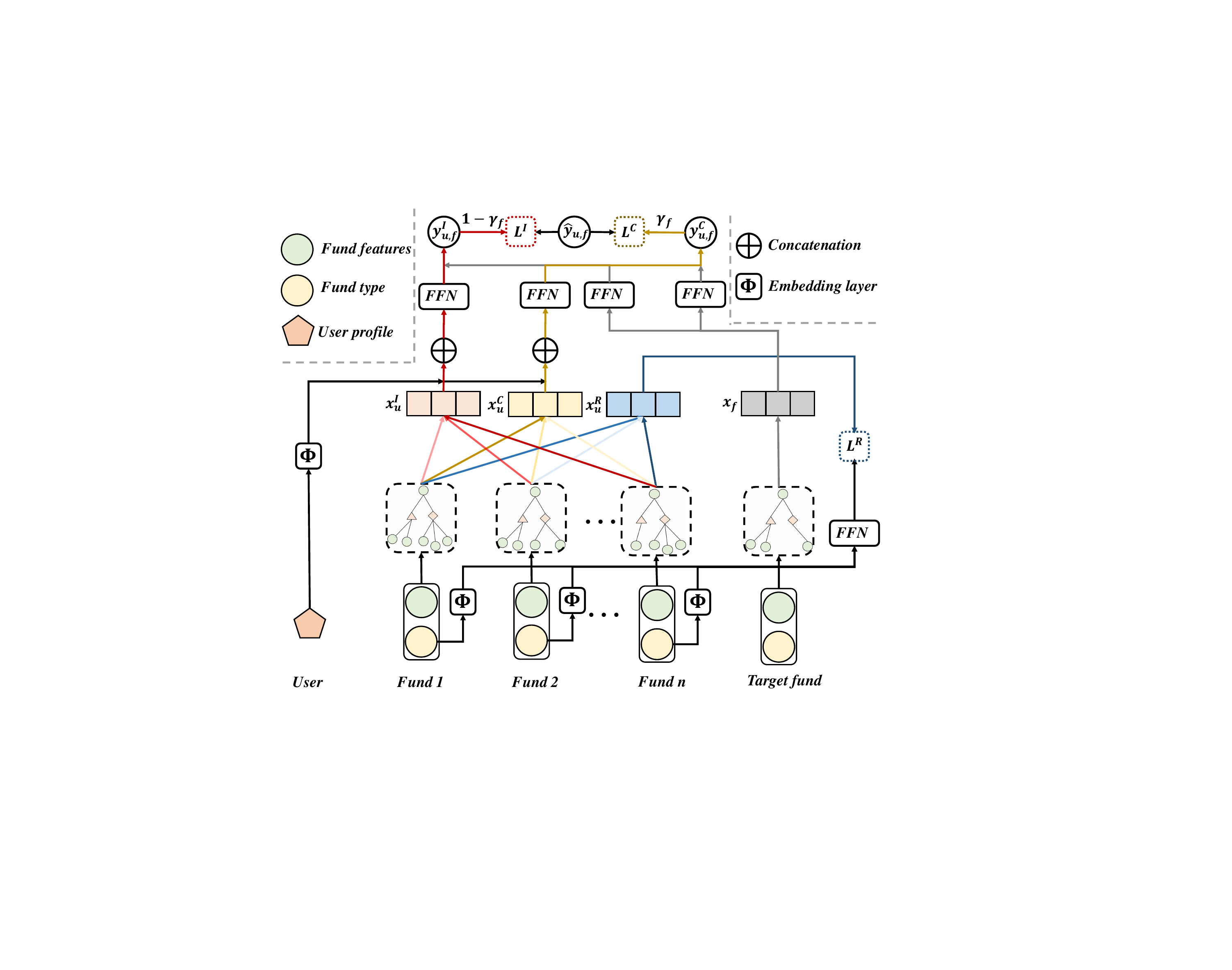}
\caption{Overall architecture of the proposed {\model}.}
\label{fig:model}
% \vspace{-2em}
% \vspace{-2.0em}
\end{figure}

In this section, we present {\model}, for intelligent matching of fund investment products, as shown in Fig.~\ref{fig:model}.

\paratitle{Incorporating Fund Graph Learning into Disentanglement.}
Actually, disentangled learning has been widely applied in traditional recommendation scenarios for multi-interest extraction~\cite{li2019multi,cen2020controllable,tian2022multi}, which could be viewed as a soft clustering process between historical behaviors. As a promising way, 
the message passing procedure of GNNs could enlarge the similarities of neighbor funds in the  graph~\cite{xia2022progcl}, and thus potentially facilitating such a clustering process. On the other hand, financial products in practical platforms essentially form a graph in nature, connected via common organizations, fund managers, types and heavyweight stocks.

i) \emph{Fund Graph Learning.}
We 
briefly 
review the well-established fund graph, 
which potentially enhances the fund representations 
and
the 
following 
disentangled process. 
We 
consider 
five relations: 
a) \underline{$\textit{fund}  \overset{manage}{\longleftrightarrow} \textit{fund manager}$}, 
b) \underline{$\textit{fund} \overset{belong\ to}{\longleftrightarrow}  \textit{organization}$},  
c) \underline{$\textit{fund} \overset{heavyweight} {\longleftrightarrow} \textit{stock}$},
d) \underline{$\textit{fund} \overset{track}{\longleftrightarrow}  \textit{stock\ index}$},
e) \underline{$\textit{fund}  
\overset{belong\ to}{\longleftrightarrow}$}
\underline{{$\textit{type}$}}.
The fund graph serves as a bridge between non-adjacent funds in historical behaviors with external knowledge for enhancing fund representations in the following disentanglement. 

Given the fund graph $\mathcal{G} = \{\mathcal{E}, \mathcal{R}\}$ with the entity set $\mathcal{E}$ and the relation set $\mathcal{R}$, following the common practice, we perform graph convolution operation $\text{Conv}(\mathcal{G}; \Theta)$ to summarize the fund graph structural information. Note that the above operation could be easily implemented as an attention~\cite{velivckovic2017graph} or a SAGE~\cite{hamilton2017inductive} convolution.

ii) \emph{Multi-granularity Representation Learning with Disentanglement.}
After extracting graph enhanced fund representation $\mathbf{H}^{(L)} \in \mathbb{R}^{|\mathcal{E}| \times d}$ with $\text{Conv}(\mathcal{G}; \Theta)$, given target user $u$'s historical behaviors $S = \left\{ {f_1,\cdots\,f_{|\mathcal{S}|}} \right\}$, we retrieve corresponding fund representations to express user’s behavior sequence as $\mathbf{X}^{S}_u \in \mathbb{R}^{|\mathcal{S}| \times d}$. 
Next, we employ the self-attention mechanism to perform disentanglement with the $d$-dimensional vector set $\{\bm{w}^{\mathcal{I}}, \bm{w}^{\mathcal{R}} , \bm{w}^{\mathcal{C}}\}$ that focus on different aspects (\ie personal \underline{I}nterest, \underline{R}isk preference and \underline{C}onformity).
\begin{equation}
\begin{split}
        \hat{\bm{\beta}}_u &= \sigma(\mathbf{X}^{S}_u\mathbf{W}^D), \\
        \{\bm{\beta}^{\mathcal{I}}_u, \bm{\beta}^{\mathcal{R}}_u, \bm{\beta}^{\mathcal{C}}_u\} &= \{\hat{\bm{\beta}}_u{\bm{w}^{\mathcal{I}}},  \hat{\bm{\beta}}_u{\bm{w}^{\mathcal{R}}}, \hat{\bm{\beta}}_u{\bm{w}^{\mathcal{C}}}\}, \\ 
        \{\mathbf{x}_u^{\mathcal{I}}, \mathbf{x}_u^{\mathcal{R}}, \mathbf{x}_u^{\mathcal{C}}\} &= \{ {\mathbf{X}^{S}_u}^\top f(\bm{\beta}^{\mathcal{I}}_u), {\mathbf{X}^{S}_u}^\top f(\bm{\beta}^{\mathcal{R}}_u), {\mathbf{X}^{S}_u}^\top f(\bm{\beta}^{\mathcal{C}}_u)\}.
 \end{split}
\end{equation}
Here, $\sigma(\cdot)$ is a non-linear function, $f(\cdot)$ is the softmax function and $\mathbf{W}^D \in \mathbb{R}^{d \times d}$ is the base weight matrix.
Although the above self-attention model has a strong capability of separating multiple aspects of user representations, disentanglement among them is not guaranteed in such an unsupervised manner~\cite{locatello2019challenging}.

\paratitle{Supervising Risk Preference with Fund Type based Contrasts.}
In fact, the entire historical behaviors related to funds provide a holistic view of user risk preference. On the other hand, we notice that the fund type is a vital factor for characterizing the risk level of funds. In light of these observations, we can abstract useful priors for risk preference from the historical fund type sequences to supervise the representation of risk preference. Formally, we denote the  historical fund type sequence of user $u$ as $\mathcal{S}^{\mathcal{T}}_u = \{t_1, \cdots, t_{|\mathcal{S}^{\mathcal{T}}_u|} \}$, and then we calculate the unifying representation of the entire interaction history as the self-supervised signal for risk preference.
\begin{equation}
    \mathbf{x}^{\mathcal{T}}_u = \text{FFN}(g(\{\Phi(t) | t \in \mathcal{S}^{\mathcal{T}}_u \})),
\end{equation}
where $\Phi(\cdot)$ denotes the ``Embedding'' operation, $g(\cdot)$ is the pooling function and $\text{FFN}(\cdot)$ represents the feed forward neural networks. 

Inspired by the  
success of contrastive learning in various applications~\cite{jaiswal2020survey}, 
we construct our self-supervised loss as follows,
\begin{equation}
\begin{split}
\mathcal{L}^{\mathcal{R}} = - \sum_{\mathcal{B}}\sum_{u \in \mathcal{B}} log\frac{exp(sim(\mathbf{x}_u^{\mathcal{R}},\mathbf{x}^{\mathcal{T}}_u)/\tau)}{\sum_{u' \sim P^{\mathcal{B}}_{neg} } exp(sim(\mathbf{x}_u^{\mathcal{R}}, \mathbf{x}_{u^{'}}^{\mathcal{T}}) / \tau)} \\
- \sum_{\mathcal{B}}\sum_{u \in \mathcal{B}} log\frac{exp(\mathbf{x}^{\mathcal{T}}_u,\mathbf{x}_u^{\mathcal{R}})/\tau)}{\sum_{u' \sim P^{\mathcal{B}}_{neg}} exp(sim(\mathbf{x}^{\mathcal{T}}_u, \mathbf{x}_{u^{'}}^{\mathcal{R}}) / \tau)},
\end{split}
\label{eq:4}
\end{equation}
where $\tau$ is the temperature parameter, and negative samples are drawn from the uniform distribution $P^{\mathcal{B}}_{neg}$ under batch $\mathcal{B}$.

\paratitle{Supervising Conformity with Fund Popularity.}
Actually, conformity encourages users with limited financial knowledge to pick popular funds, which are always highly recommended by fund managers and even the public. Hence, it inspires that the \emph{fund popularity} is a critical factor to capture conformity. Formally, we define the popularity of target fund $f$ as follows,
\begin{equation}
\gamma_f = \frac{\log{C_f} - \log{C_{min}}}{\log{C_{max}} - \log{C_{min}}}.
% \gamma_f = {(\log{C_f} - \log{C_{min}})} / {(\log{C_{max}} - \log{C_{min}})}.
\label{eq:5}
\end{equation}
Here, $C_f$ denotes the number of user interactions \wrt fund $f$  while $C_{max} = \max_{f \in \mathcal{F}}{C_f}$ and $C_{min} = \min_{f \in \mathcal{F}}{C_f}$ respectively represent the maximum and the minimum, where $\mathcal{F}$ is the fund set. 
Meanwhile, given target user $u$ and fund $f$, we can obtain the conformity based score as follows,
\begin{equation}
        y_{u,f}^{\mathcal{C}} = \sigma(\text{FFN}^{\mathcal{C}}(\mathbf{x}_u^{\mathcal{P}}||\mathbf{x}_u^{\mathcal{C}})^\top \cdot \text{FFN}^{\mathcal{C}}(\mathbf{x}_f)),
\end{equation}
where $\mathbf{x}_u^{\mathcal{P}}$ is the feature vector of user basic profile, $\mathbf{x}_f$ is the fund representation retrieved from $\mathbf{H}^{(L)}$, ``$||$'' is the concatenation operation and $\sigma(\cdot)$ is the sigmoid function. Considering the positive correlation between conformity score and fund popularity, we formulate the conformity-side loss function in the following supervised way,
\begin{equation}
    \mathcal{L}^\mathcal{C} = \gamma_f \cdot \text{C-E}(y^\mathcal{C}_{u,f}, \hat{y}_{u, f}),
\end{equation}
where $\hat{y}_{u, f}$ is the ground truth and $\text{C-E}(\cdot)$ represents the cross entropy loss.
Analogously, personal interest can be 
modelled in the above similar way where funds with low popularity are the core. 
\begin{equation}
    \begin{split}
        y_{u,f}^{\mathcal{I}} &= \sigma(\text{FFN}^{\mathcal{I}}(\mathbf{x}_u^{\mathcal{P}}||\mathbf{x}_u^{\mathcal{I}})^\top \cdot \text{FFN}^{\mathcal{I}}(\mathbf{x}_f)), \\
         \mathcal{L}^\mathcal{I} &= (1 - \gamma_f) \cdot \text{C-E}(y^\mathcal{I}_{u,f}, \hat{y}_{u, f}).
    \end{split}
\end{equation}

\paratitle{Putting All Together and Making Prediction.}
By integrating all the above loss functions, the overall objective function for the proposed {\model} is defined as follows,
\begin{equation}
\mathcal{L} = \mathcal{L}^{\mathcal{I}} + \mathcal{L}^{\mathcal{C}} + \epsilon \cdot \mathcal{L}^{\mathcal{R}},
\label{eq:8}
\end{equation}
where $\epsilon \geq 0$ controls the risk preference term $\mathcal{L}^{\mathcal{R}}$. 
At last, {\model} considers both conformity and interest for the final prediction, 
\begin{equation}
    y_{u,f} = \gamma_f \cdot y_{u,f}^{\mathcal{C}} + (1 - \gamma_f) \cdot y_{u,f}^{\mathcal{I}}.
\end{equation}
\section{Experiments}

% In this section, we conduct extensive experimental evaluation on four industrial datasets, and then present comprehensive result analysis.

% \begin{table}[H]
% \setlength{\tabcolsep}{1.3mm}{
% \begin{tabular}{ccccc}
% \toprule
% {} &{Jan.} & {Feb.} & {Mar.}  & {Apr.}\\
% \midrule
% {\# User}  & {9,398,526} & {10,554,837} & {10,239,927} & {9,978,665}\\
% {\# Fund}  & {10,140} & {10,046} & {10,030} & {10,074} \\
% \midrule
% {\# Train}   & {4,419,505}& {5,835,280}& {5,431,277} & {3,909,320}\\
% {\# Val.}    & {54,684}& {55,193}& {43,118} & {54,681}\\
% {\# Test}    & {8,168,575}& {8,187,197}& {8,160,572}& {8,693,234} \\
% \bottomrule
% \end{tabular}}
% \caption{\textcolor{red}{Statistics of the datasets in the experimental evaluation.} \label{tab:off_data}}
% \end{table}

% % \vspace{-2.0em}
% \begin{figure}[H]
% % \vspace{-1.5em}
% 	\centering
% 	\subfigure[Jan.]{\includegraphics[width=0.45\columnwidth]{figures/graph_ndcg_A.pdf}}
% 	\subfigure[Feb.]{\includegraphics[width=0.45\columnwidth]{figures/graph_ndcg_c.pdf}}
  
% % 	\subfigure[Jan. - NDCG]{\includegraphics[width=0.45\columnwidth]{figures/graph_ndcg_A.pdf}}
% %   \subfigure[Feb. -  NDCG]{\includegraphics[width=0.45\columnwidth]{figures/graph_ndcg_c.pdf}}
%   \caption{Impact of fund graph learning.}
% 	\label{fig:ab_study_graph}
% % 	\vspace{-2em}
% \end{figure}

% \vspace{-1.0em}
\begin{table*}[t]\centering
\setlength{\tabcolsep}{1.6mm}{
\begin{tabular}{c|c|c|c|c|c||c|c|c|c}
\toprule
% \multirow{2}{*}{Datasets} 
% & \multirow{2}{*}{Methods} 
% & \multicolumn{4}{c|}{Recall@K} 
% & \multicolumn{4}{c}{NDCG@K} \\   
% \cline{3-10}
%\midrule
{Datasets} & {Methods} & {Recall@5} & {Recall@10} & {Recall@15} & {Recall@20} & {NDCG@5} &{NDCG@10} & {NDCG@15} & {NDCG@20}  \\
\hline 
{} & {SASRec~\cite{kang2018self}} 
% & {0.9564} 
& {0.1805} & {0.2537} & {0.2985} & {0.3313} & {0.1192} & {0.1428} & {0.1547} & {0.1624} \\
{} & {ComiRec~\cite{cen2020controllable}} 
% & {0.9371} 
& {0.1324} & {0.1846} & {0.2185} & {0.2519} & {0.0813} & {0.0978} & {0.1068} & {0.1147} \\
{Jan.} & {LightGCN~\cite{he2020lightgcn}} 
% & {0.9306} 
& {0.1148} & {0.1755} & {0.2068} & {0.2330} & {0.0729} & {0.0927} & {0.1010} & {0.1071} \\
{} & {DisenGCN~\cite{ma2019disentangled,zhuang2020hubble}} 
% & {0.9407} 
& {0.1363} & {0.1869} & {0.2240} & {0.2545} & {0.0924} & {0.1086} & {0.1184} & {0.1256} \\
{} & {{\model}} 
% & {\textbf{0.9616}} 
& {\textbf{0.2088}} & {\textbf{0.2892}} & {\textbf{0.3338}} & {\textbf{0.3680}} & {\textbf{0.1424}} & {\textbf{0.1686}} & {\textbf{0.1804}} & {\textbf{0.1885}}\\

\hline 
{} & {SASRec~\cite{kang2018self}} 
% & {0.9603} 
& {0.1861} & {0.2471} & {0.2910} & {0.3251} & {0.1253} & {0.1450} & {0.1565} & {0.1646} \\
{} & {ComiRec~\cite{cen2020controllable}}  
% & {0.9379} 
& {0.1282} & {0.1830} & {0.2306} & {0.2556} & {0.0838} & {0.1014} & {0.1140} & {0.1199}\\
{Feb.} & {LightGCN~\cite{he2020lightgcn}} 
% & {0.9393} 
& {0.1399} & {0.1932} & {0.2281} & {0.2566} & {0.0891} & {0.1063} & {0.1156} & {0.1223} \\
{} & {DisenGCN~\cite{ma2019disentangled,zhuang2020hubble}}  
% & {0.9435} 
& {0.1389} & {0.2017} & {0.2369} & {0.2630} & {0.0866} & {0.1070} & {0.1163} & {0.1224} \\
{} & {{\model}}  
% & {\textbf{0.9623}} 
& {\textbf{0.2069}} & {\textbf{0.2752}} & {\textbf{0.3188}} & {\textbf{0.3514}} & {\textbf{0.1424}} & {\textbf{0.1644}} & {\textbf{0.1760}} & {\textbf{0.1837}}\\

\hline 
{} & {SASRec~\cite{kang2018self}} 
% & {0.9688} 
& {0.2054} & {0.2720} & {0.3138} & {0.3480} & {0.1489} & {0.1703} & {0.1814} & {0.1895} \\
{} & {ComiRec~\cite{cen2020controllable}} 
% & {0.9457} 
& {0.1231} & {0.1840} & {0.2165} & {0.2438} & {0.0802} & {0.0998} & {0.1085} & {0.1149}\\
{Mar.} & {LightGCN~\cite{he2020lightgcn}}  
% & {0.9517}
& {0.1258} & {0.1767} & {0.2173} & {0.2533} & {0.0856} & {0.1019} & {0.1126} & {0.1211} \\
{} & {DisenGCN~\cite{ma2019disentangled,zhuang2020hubble}} 
% & {0.9595} 
& {0.1441} & {0.2068} & {0.2536} & {0.2895} & {0.0934} & {0.1136} & {0.1260} & {0.1344} \\
{} & {{\model}} 
% & {\textbf{0.9705}} 
& {\textbf{0.2423}} & {\textbf{0.3131}} & {\textbf{0.3591}} & {\textbf{0.3935}} & {\textbf{0.1646}} & {\textbf{0.1875}} & {\textbf{0.1997}} & {\textbf{0.2078}}\\

\hline 
{} & {SASRec~\cite{kang2018self}} 
% & {0.9627} 
& {0.2113} & {0.2734} & {0.3110} & {0.3380} & {0.1452} & {0.1653} & {0.1752} & {0.1816} \\
{} & {ComiRec~\cite{cen2020controllable}} 
% & {0.9289} 
& {0.1129} & {0.1871} & {0.2204} & {0.2434} & {0.0809} & {0.1042} & {0.1130} & {0.1184}\\
{Apr.} & {LightGCN~\cite{he2020lightgcn}}  
% & {0.9464} 
& {0.1243} & {0.1782} & {0.2130} & {0.2402} & {0.0814} & {0.0989} & {0.1081} & {0.1145} \\
{} & {DisenGCN~\cite{ma2019disentangled,zhuang2020hubble}}  
% & {0.9528} 
& {0.1607} & {0.2192} & {0.2543} & {0.2836} & {0.1056} & {0.1247} & {0.1340} & {0.1409} \\
{} & {{\model}}  
% & {\textbf{0.9721}} 
& {\textbf{0.2295}} & {\textbf{0.2924}} & {\textbf{0.3313}} & {\textbf{0.3614}} & {\textbf{0.1636}} & {\textbf{0.1839}} & {\textbf{0.1942}} & {\textbf{0.2014}}\\

\bottomrule

\end{tabular}}
\caption{Overall performance evaluation across four offline datasets. The best results are highlighted in boldface.} 
\label{tab:overall_result}
\end{table*}
% \vspace{-2.0em}

\paratitle{Dataset Description.}
We collect a real-world large-scale dataset\footnote{The dataset does not contain any Personalized Identifiable Information.} from one of the biggest financial platforms in China, and 
extract four sub-datasets by month for performance evaluation, namely \textbf{Jan.}, \textbf{Feb.}, \textbf{Mar.} and \textbf{Apr.}. 
Specifically, 
for each month, we leave out interactions on the last day as the test set and utilize the remaining data for training.
Moreover, we hold out a part of the training data as the validation set for parameter tuning.
Due to the huge volume of real-world interaction records, the daily sampling strategy is applied in each sub-dataset.
Finally, each sub-dataset includes about \textbf{one million} users and about \textbf{ten thousand} funds, with about \textbf{fifty million} records for training, about \textbf{half a million}  records for validation and about \textbf{eight million} records for testing. Meanwhile, we organize the fund graph with about \textbf{ten thousand} entities and about \textbf{half a million} relations.

\begin{figure}
% \vspace{-1.25em}
	\centering
	\subfigure[Jan. - Ablation study I]{\includegraphics[width=0.45\columnwidth]{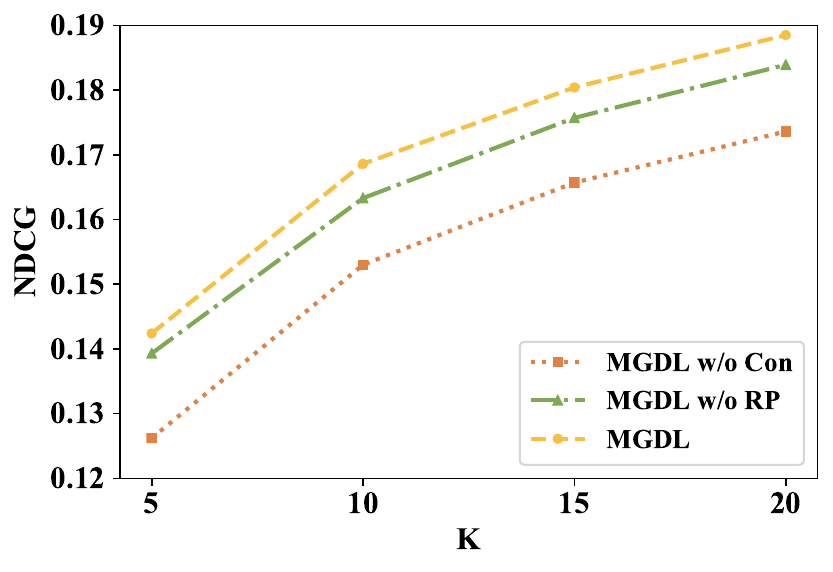}}
	\subfigure[Feb. - Ablation study I]{\includegraphics[width=0.45\columnwidth]{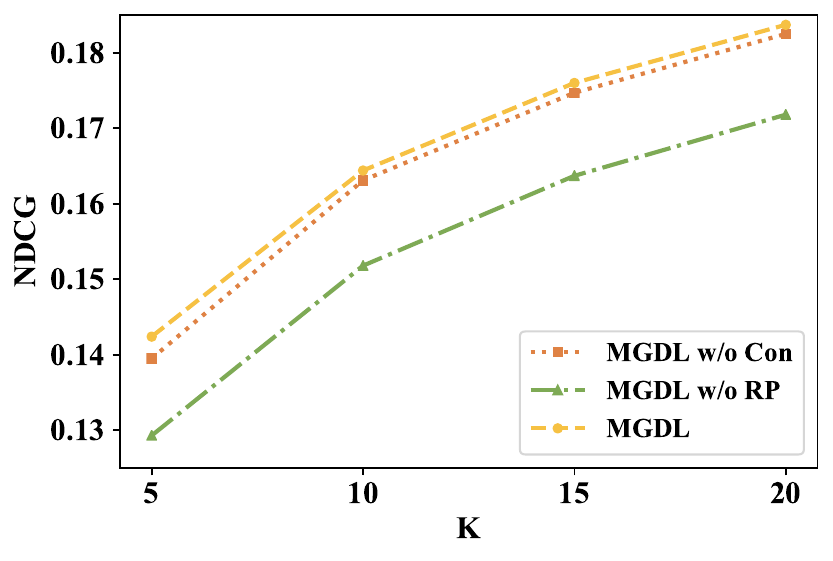}}
  
 	\subfigure[Jan. - Ablation study II]{\includegraphics[width=0.45\columnwidth]{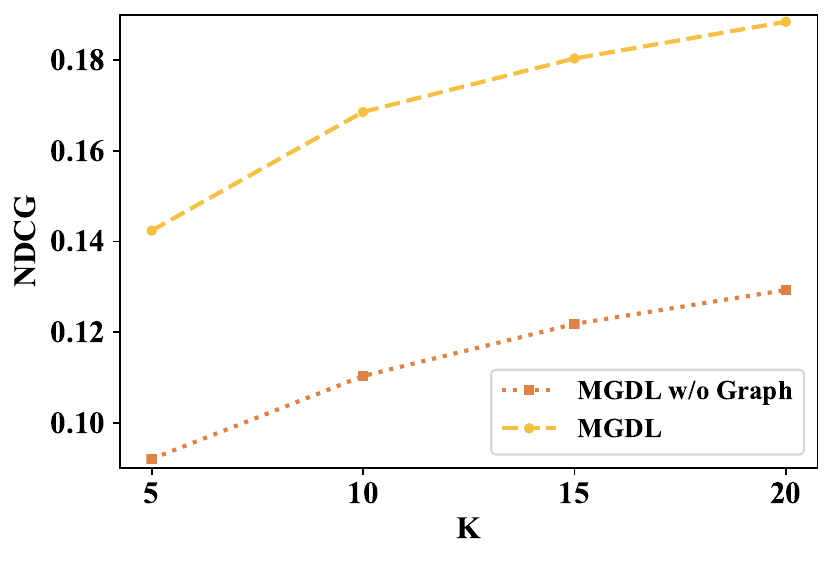}}
  \subfigure[Feb. - Ablation study II]{\includegraphics[width=0.45\columnwidth]{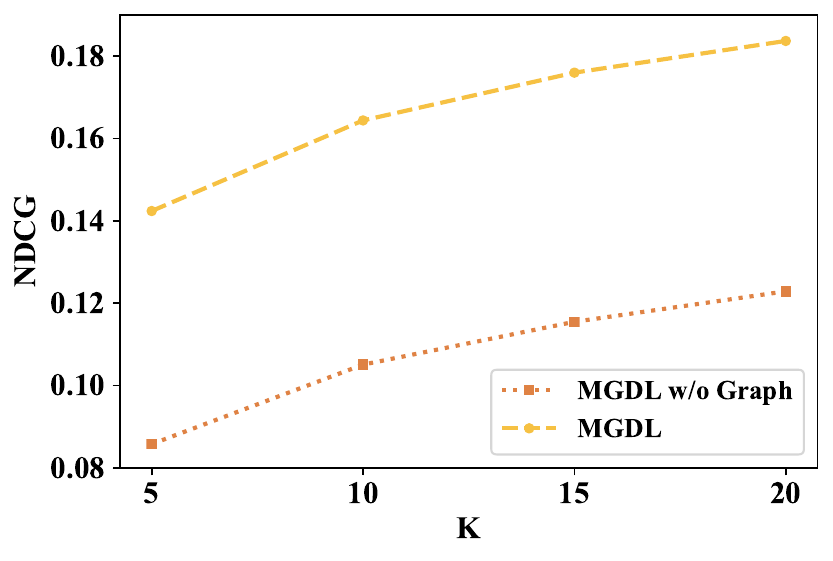}}
  \caption{Ablation studies \wrt NDCG. Similar trends could also be observed on Mar. and Apr. datasets.}
	\label{fig:ab_study}
	% \vspace{-2em}
\end{figure}

\paratitle{Overall Performance.}
We report the overall comparison results in Table~\ref{tab:overall_result}.
Note that the fund graph is adopted in {\model}, thus we extend LightGCN and DisenGCN to adapt to the mixed graph consisting of the user-item bipartite graph and the fund graph for a fair comparison. Besides, we find NGCF~\cite{wang2019neural} , KGAT~\cite{wang2019kgat} and DGCF~\cite{wang2020disentangled} achieve relatively poor performance when compared to above selected baselines, and thus we omit them in our experimental results.
We find that {\model} outperforms all baselines by a large margin in all cases, indicating the superiority of supplementing the fund recommendation issue with both conformity and risk preference modelling via  the  multi-granularity graph disentangled learning. Moreover, the performance gain of DisenGCN \wrt ComiRec reveals the usefulness of fund graph structure for pulling similar funds closer in the disentangled process, while SASRec works remarkably well among these baselines, intuitively attributed to the powerful ability of Transformer architecture.

\paratitle{Ablation I: Impact of Multi-granularity Disentangled Learning.}
We prepare two variants of {\model}, namely 
i) \underline{\textbf{{\model} w/o Con}}, which removes the conformity part and 
ii) \underline{\textbf{{\model} w/o RP}}, which removes the risk preference modelling.
From Fig.~\ref{fig:ab_study} (a) and (b) we observe that the complete {\model} achieves the best performance in all cases across evaluation metrics. 
It indicates that both conformity and risk preference are indispensable to the fund recommendation task, and the well-designed disentangled component with self-supervision endows {\model} with more meaningful representations.

\paratitle{Ablation II: Effectiveness Analysis of Fund Graph Learning.}
Next, we zoom into the effectiveness of  the fund graph learning towards {\model}, and specifically denote the variant removing the fund graph learning as \underline{\textbf{{\model} w/o Graph}}. Not surprisingly, we observe that the performance of {\model} drops a lot without fund graph learning in Fig.~\ref{fig:ab_study} (c) and (d), revealing that the fund graph structure, as a critical prior, could greatly contribute to {\model}.

\paratitle{Visualization Analysis.}
To examine the capability of {\model} intuitively, 
we visualize the conformity- and personal interest-side user representations (\ie $\mathbf{x}_u^{\mathcal{C}}$ and $\mathbf{x}_u^{\mathcal{I}}$) using $t$-SNE, since they are used for the final predictions. 
We label each user according to his/her fund holding level: 0$\sim$4 for $\mathbf{x}_u^{\mathcal{C}}$ and 5$\sim$9 for $\mathbf{x}_u^{\mathcal{I}}$, \eg users hold 0$\sim$100 in our platform would be labeled as 0 for $\mathbf{x}_u^{\mathcal{C}}$ and 5 for $\mathbf{x}_u^{\mathcal{I}}$.

\begin{figure}[H]
	\centering
	\subfigure[]{\includegraphics[width=0.44\columnwidth]{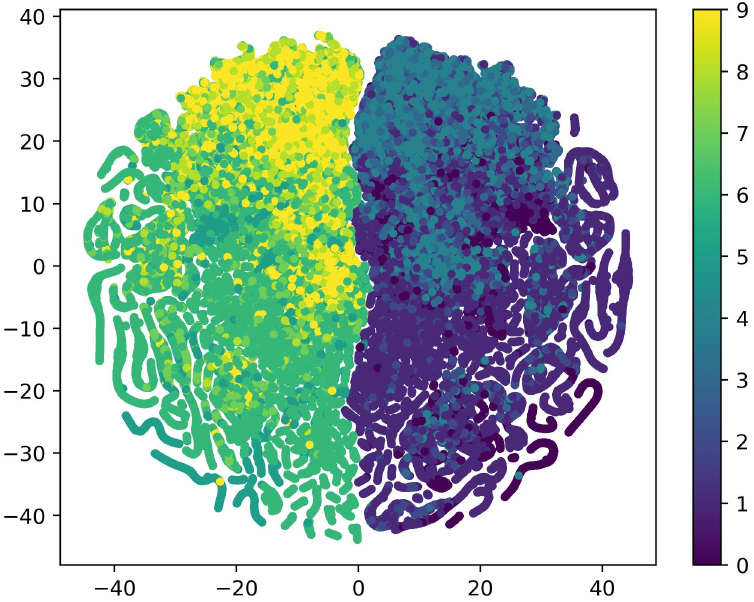}}
	\subfigure[]{\includegraphics[width=0.44\columnwidth]{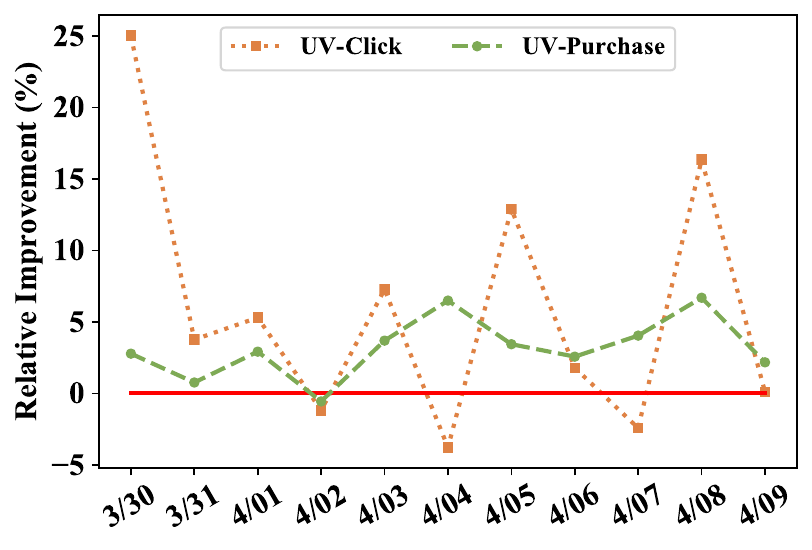}}
  
  \caption{(a) Visualization of predictive user embeddings learned by {\model}. (b) Online performance.}
	\label{fig:vis_online}
% 	\vspace{-2.0em}
\end{figure}

From Fig.~\ref{fig:vis_online} (a), 
we find that:
i) {\model} can reasonably separate the conformity- and personal interest-side representations and 
learn a relatively crisp boundary.
It  depicts  that  user conformity is well distinguished by {\model} through our proposed self-supervised signal, \ie fund  popularity.
ii) Both of the conformity- and personal interest-side representations are well layered \wrt the user holding level, which shows that {\model} could well reflect the risk preference 
even though no relevant label (\ie user holding level) is available.

% \vspace{-2.0em}

\paratitle{Online Performance.}
We  deploy {\model} on our platform for A/B test against a baseline
model based on FM and Transformer.
We perform 
online evaluation from ``2022/3/30'' to ``2022/4/09'' 
via
metrics \textbf{UV-Click} and \textbf{UV-Purchase}~\footnote{UV means unique visitor}, and show the experimental results in Fig.~\ref{fig:vis_online} (b). Compared to the 
baseline (\ie the red solid line in the Fig.~\ref{fig:vis_online} (b)), {\model} gains the overall improvements of 3.12\% and 6.92\% \wrt UV-Click and UV-Purchase, 
which are both statistically significant with a significance level of 95\%. This practice-oriented experiment 
further 
demonstrates the superiority of {\model}.

\section{Conclusion}

In this paper, we propose {\model} to perform effective intelligent matching of fund investment products, where both conformity and risk preference are emphasized in making fund investment decisions beyond personal interest.  
Comprehensive experiments in offline/online environments demonstrate the superiority of {\model}.

\bibliographystyle{ACM-Reference-Format}
\bibliography{sample-base}

%%% -*-BibTeX-*-
%%% Do NOT edit. File created by BibTeX with style
%%% ACM-Reference-Format-Journals [18-Jan-2012].

\begin{thebibliography}{28}

%%% ====================================================================
%%% NOTE TO THE USER: you can override these defaults by providing
%%% customized versions of any of these macros before the \bibliography
%%% command.  Each of them MUST provide its own final punctuation,
%%% except for \shownote{}, \showDOI{}, and \showURL{}.  The latter two
%%% do not use final punctuation, in order to avoid confusing it with
%%% the Web address.
%%%
%%% To suppress output of a particular field, define its macro to expand
%%% to an empty string, or better, \unskip, like this:
%%%
%%% \newcommand{\showDOI}[1]{\unskip}   % LaTeX syntax
%%%
%%% \def \showDOI #1{\unskip}           % plain TeX syntax
%%%
%%% ====================================================================

\ifx \showCODEN    \undefined \def \showCODEN     #1{\unskip}     \fi
\ifx \showDOI      \undefined \def \showDOI       #1{#1}\fi
\ifx \showISBNx    \undefined \def \showISBNx     #1{\unskip}     \fi
\ifx \showISBNxiii \undefined \def \showISBNxiii  #1{\unskip}     \fi
\ifx \showISSN     \undefined \def \showISSN      #1{\unskip}     \fi
\ifx \showLCCN     \undefined \def \showLCCN      #1{\unskip}     \fi
\ifx \shownote     \undefined \def \shownote      #1{#1}          \fi
\ifx \showarticletitle \undefined \def \showarticletitle #1{#1}   \fi
\ifx \showURL      \undefined \def \showURL       {\relax}        \fi
% The following commands are used for tagged output and should be
% invisible to TeX
\providecommand\bibfield[2]{#2}
\providecommand\bibinfo[2]{#2}
\providecommand\natexlab[1]{#1}
\providecommand\showeprint[2][]{arXiv:#2}

\bibitem[\protect\citeauthoryear{Cen, Zhang, Zou, Zhou, Yang, and Tang}{Cen
  et~al\mbox{.}}{2020}]%
        {cen2020controllable}
\bibfield{author}{\bibinfo{person}{Yukuo Cen}, \bibinfo{person}{Jianwei Zhang},
  \bibinfo{person}{Xu Zou}, \bibinfo{person}{Chang Zhou},
  \bibinfo{person}{Hongxia Yang}, {and} \bibinfo{person}{Jie Tang}.}
  \bibinfo{year}{2020}\natexlab{}.
\newblock \showarticletitle{Controllable multi-interest framework for
  recommendation}. In \bibinfo{booktitle}{\emph{SIGKDD}}.
  \bibinfo{pages}{2942--2951}.
\newblock


\bibitem[\protect\citeauthoryear{Fan, Liu, Jin, Zhao, Tang, and Li}{Fan
  et~al\mbox{.}}{2022}]%
        {fan2022graph}
\bibfield{author}{\bibinfo{person}{Wenqi Fan}, \bibinfo{person}{Xiaorui Liu},
  \bibinfo{person}{Wei Jin}, \bibinfo{person}{Xiangyu Zhao},
  \bibinfo{person}{Jiliang Tang}, {and} \bibinfo{person}{Qing Li}.}
  \bibinfo{year}{2022}\natexlab{}.
\newblock \showarticletitle{Graph Trend Filtering Networks for Recommendation}.
  In \bibinfo{booktitle}{\emph{SIGIR}}. \bibinfo{pages}{112--121}.
\newblock


\bibitem[\protect\citeauthoryear{Fan, Ma, Li, He, Zhao, Tang, and Yin}{Fan
  et~al\mbox{.}}{2019}]%
        {fan2019graph}
\bibfield{author}{\bibinfo{person}{Wenqi Fan}, \bibinfo{person}{Yao Ma},
  \bibinfo{person}{Qing Li}, \bibinfo{person}{Yuan He}, \bibinfo{person}{Eric
  Zhao}, \bibinfo{person}{Jiliang Tang}, {and} \bibinfo{person}{Dawei Yin}.}
  \bibinfo{year}{2019}\natexlab{}.
\newblock \showarticletitle{Graph neural networks for social recommendation}.
  In \bibinfo{booktitle}{\emph{WWW}}. \bibinfo{pages}{417--426}.
\newblock


\bibitem[\protect\citeauthoryear{Fan, Liu, Lian, Zhao, Xie, and Wen}{Fan
  et~al\mbox{.}}{2021}]%
        {fan2021lighter}
\bibfield{author}{\bibinfo{person}{Xinyan Fan}, \bibinfo{person}{Zheng Liu},
  \bibinfo{person}{Jianxun Lian}, \bibinfo{person}{Wayne~Xin Zhao},
  \bibinfo{person}{Xing Xie}, {and} \bibinfo{person}{Ji-Rong Wen}.}
  \bibinfo{year}{2021}\natexlab{}.
\newblock \showarticletitle{Lighter and better: low-rank decomposed
  self-attention networks for next-item recommendation}. In
  \bibinfo{booktitle}{\emph{SIGIR}}. \bibinfo{pages}{1733--1737}.
\newblock


\bibitem[\protect\citeauthoryear{Hamilton, Ying, and Leskovec}{Hamilton
  et~al\mbox{.}}{2017}]%
        {hamilton2017inductive}
\bibfield{author}{\bibinfo{person}{Will Hamilton}, \bibinfo{person}{Zhitao
  Ying}, {and} \bibinfo{person}{Jure Leskovec}.}
  \bibinfo{year}{2017}\natexlab{}.
\newblock \showarticletitle{Inductive representation learning on large graphs}.
  In \bibinfo{booktitle}{\emph{NIPS}}. \bibinfo{pages}{1024--1034}.
\newblock


\bibitem[\protect\citeauthoryear{He, Deng, Wang, Li, Zhang, and Wang}{He
  et~al\mbox{.}}{2020}]%
        {he2020lightgcn}
\bibfield{author}{\bibinfo{person}{Xiangnan He}, \bibinfo{person}{Kuan Deng},
  \bibinfo{person}{Xiang Wang}, \bibinfo{person}{Yan Li},
  \bibinfo{person}{Yongdong Zhang}, {and} \bibinfo{person}{Meng Wang}.}
  \bibinfo{year}{2020}\natexlab{}.
\newblock \showarticletitle{Lightgcn: Simplifying and powering graph
  convolution network for recommendation}. In
  \bibinfo{booktitle}{\emph{SIGIR}}. \bibinfo{pages}{639--648}.
\newblock


\bibitem[\protect\citeauthoryear{Hou, Hu, Zhang, and Zhao}{Hou
  et~al\mbox{.}}{2022}]%
        {hou2022core}
\bibfield{author}{\bibinfo{person}{Yupeng Hou}, \bibinfo{person}{Binbin Hu},
  \bibinfo{person}{Zhiqiang Zhang}, {and} \bibinfo{person}{Wayne~Xin Zhao}.}
  \bibinfo{year}{2022}\natexlab{}.
\newblock \showarticletitle{Core: simple and effective session-based
  recommendation within consistent representation space}. In
  \bibinfo{booktitle}{\emph{SIGIR}}. \bibinfo{pages}{1796--1801}.
\newblock


\bibitem[\protect\citeauthoryear{Hu, Shi, Zhao, and Yu}{Hu
  et~al\mbox{.}}{2018}]%
        {hu2018leveraging}
\bibfield{author}{\bibinfo{person}{Binbin Hu}, \bibinfo{person}{Chuan Shi},
  \bibinfo{person}{Wayne~Xin Zhao}, {and} \bibinfo{person}{Philip~S Yu}.}
  \bibinfo{year}{2018}\natexlab{}.
\newblock \showarticletitle{Leveraging meta-path based context for top-n
  recommendation with a neural co-attention model}. In
  \bibinfo{booktitle}{\emph{SIGKDD}}. \bibinfo{pages}{1531--1540}.
\newblock


\bibitem[\protect\citeauthoryear{Jaiswal, Babu, Zadeh, Banerjee, and
  Makedon}{Jaiswal et~al\mbox{.}}{2020}]%
        {jaiswal2020survey}
\bibfield{author}{\bibinfo{person}{Ashish Jaiswal},
  \bibinfo{person}{Ashwin~Ramesh Babu}, \bibinfo{person}{Mohammad~Zaki Zadeh},
  \bibinfo{person}{Debapriya Banerjee}, {and} \bibinfo{person}{Fillia
  Makedon}.} \bibinfo{year}{2020}\natexlab{}.
\newblock \showarticletitle{A survey on contrastive self-supervised learning}.
  In \bibinfo{booktitle}{\emph{Technologies}}, Vol.~\bibinfo{volume}{9}.
  \bibinfo{pages}{2}.
\newblock


\bibitem[\protect\citeauthoryear{Kang and McAuley}{Kang and McAuley}{2018}]%
        {kang2018self}
\bibfield{author}{\bibinfo{person}{Wang-Cheng Kang} {and}
  \bibinfo{person}{Julian McAuley}.} \bibinfo{year}{2018}\natexlab{}.
\newblock \showarticletitle{Self-attentive sequential recommendation}. In
  \bibinfo{booktitle}{\emph{ICDM}}. \bibinfo{pages}{197--206}.
\newblock


\bibitem[\protect\citeauthoryear{Li, Liu, Wu, Xu, Zhao, Huang, Kang, Chen, Li,
  and Lee}{Li et~al\mbox{.}}{2019}]%
        {li2019multi}
\bibfield{author}{\bibinfo{person}{Chao Li}, \bibinfo{person}{Zhiyuan Liu},
  \bibinfo{person}{Mengmeng Wu}, \bibinfo{person}{Yuchi Xu},
  \bibinfo{person}{Huan Zhao}, \bibinfo{person}{Pipei Huang},
  \bibinfo{person}{Guoliang Kang}, \bibinfo{person}{Qiwei Chen},
  \bibinfo{person}{Wei Li}, {and} \bibinfo{person}{Dik~Lun Lee}.}
  \bibinfo{year}{2019}\natexlab{}.
\newblock \showarticletitle{Multi-interest network with dynamic routing for
  recommendation at Tmall}. In \bibinfo{booktitle}{\emph{CIKM}}.
  \bibinfo{pages}{2615--2623}.
\newblock


\bibitem[\protect\citeauthoryear{Liu, Wu, Zhang, and Shen}{Liu
  et~al\mbox{.}}{2022}]%
        {liu2022unify}
\bibfield{author}{\bibinfo{person}{Xiaoming Liu}, \bibinfo{person}{Shaocong
  Wu}, \bibinfo{person}{Zhaohan Zhang}, {and} \bibinfo{person}{Chao Shen}.}
  \bibinfo{year}{2022}\natexlab{}.
\newblock \showarticletitle{Unify Local and Global Information for Top-N
  Recommendation}. In \bibinfo{booktitle}{\emph{SIGIR}}.
  \bibinfo{pages}{1262--1272}.
\newblock


\bibitem[\protect\citeauthoryear{Locatello, Bauer, Lucic, Raetsch, Gelly,
  Sch{\"o}lkopf, and Bachem}{Locatello et~al\mbox{.}}{2019}]%
        {locatello2019challenging}
\bibfield{author}{\bibinfo{person}{Francesco Locatello},
  \bibinfo{person}{Stefan Bauer}, \bibinfo{person}{Mario Lucic},
  \bibinfo{person}{Gunnar Raetsch}, \bibinfo{person}{Sylvain Gelly},
  \bibinfo{person}{Bernhard Sch{\"o}lkopf}, {and} \bibinfo{person}{Olivier
  Bachem}.} \bibinfo{year}{2019}\natexlab{}.
\newblock \showarticletitle{Challenging common assumptions in the unsupervised
  learning of disentangled representations}. In
  \bibinfo{booktitle}{\emph{ICML}}. \bibinfo{pages}{4114--4124}.
\newblock


\bibitem[\protect\citeauthoryear{Ma, Cui, Kuang, Wang, and Zhu}{Ma
  et~al\mbox{.}}{2019}]%
        {ma2019disentangled}
\bibfield{author}{\bibinfo{person}{Jianxin Ma}, \bibinfo{person}{Peng Cui},
  \bibinfo{person}{Kun Kuang}, \bibinfo{person}{Xin Wang}, {and}
  \bibinfo{person}{Wenwu Zhu}.} \bibinfo{year}{2019}\natexlab{}.
\newblock \showarticletitle{Disentangled graph convolutional networks}. In
  \bibinfo{booktitle}{\emph{ICML}}. \bibinfo{pages}{4212--4221}.
\newblock


\bibitem[\protect\citeauthoryear{Shan, Hoens, Jiao, Wang, Yu, and Mao}{Shan
  et~al\mbox{.}}{2016}]%
        {shan2016deep}
\bibfield{author}{\bibinfo{person}{Ying Shan}, \bibinfo{person}{T~Ryan Hoens},
  \bibinfo{person}{Jian Jiao}, \bibinfo{person}{Haijing Wang},
  \bibinfo{person}{Dong Yu}, {and} \bibinfo{person}{JC Mao}.}
  \bibinfo{year}{2016}\natexlab{}.
\newblock \showarticletitle{Deep crossing: Web-scale modeling without manually
  crafted combinatorial features}. In \bibinfo{booktitle}{\emph{SIGKDD}}.
  \bibinfo{pages}{255--262}.
\newblock


\bibitem[\protect\citeauthoryear{Shi, Hu, Zhao, and Philip}{Shi
  et~al\mbox{.}}{2018}]%
        {shi2018heterogeneous}
\bibfield{author}{\bibinfo{person}{Chuan Shi}, \bibinfo{person}{Binbin Hu},
  \bibinfo{person}{Wayne~Xin Zhao}, {and} \bibinfo{person}{S~Yu Philip}.}
  \bibinfo{year}{2018}\natexlab{}.
\newblock \showarticletitle{Heterogeneous information network embedding for
  recommendation}.
\newblock \bibinfo{journal}{\emph{IEEE Transactions on Knowledge and Data
  Engineering}} \bibinfo{volume}{31}, \bibinfo{number}{2}
  (\bibinfo{year}{2018}), \bibinfo{pages}{357--370}.
\newblock


\bibitem[\protect\citeauthoryear{Sun, Liu, Wu, Pei, Lin, Ou, and Jiang}{Sun
  et~al\mbox{.}}{2019}]%
        {sun2019bert4rec}
\bibfield{author}{\bibinfo{person}{Fei Sun}, \bibinfo{person}{Jun Liu},
  \bibinfo{person}{Jian Wu}, \bibinfo{person}{Changhua Pei},
  \bibinfo{person}{Xiao Lin}, \bibinfo{person}{Wenwu Ou}, {and}
  \bibinfo{person}{Peng Jiang}.} \bibinfo{year}{2019}\natexlab{}.
\newblock \showarticletitle{BERT4Rec: Sequential recommendation with
  bidirectional encoder representations from transformer}. In
  \bibinfo{booktitle}{\emph{CIKM}}. \bibinfo{pages}{1441--1450}.
\newblock


\bibitem[\protect\citeauthoryear{Tian, Chang, Niu, Song, and Li}{Tian
  et~al\mbox{.}}{2022}]%
        {tian2022multi}
\bibfield{author}{\bibinfo{person}{Yu Tian}, \bibinfo{person}{Jianxin Chang},
  \bibinfo{person}{Yanan Niu}, \bibinfo{person}{Yang Song}, {and}
  \bibinfo{person}{Chenliang Li}.} \bibinfo{year}{2022}\natexlab{}.
\newblock \showarticletitle{When Multi-Level Meets Multi-Interest: A
  Multi-Grained Neural Model for Sequential Recommendation}. In
  \bibinfo{booktitle}{\emph{SIGIR}}. \bibinfo{pages}{1632--1641}.
\newblock


\bibitem[\protect\citeauthoryear{Veli{\v{c}}kovi{\'c}, Cucurull, Casanova,
  Romero, Lio, and Bengio}{Veli{\v{c}}kovi{\'c} et~al\mbox{.}}{2018}]%
        {velivckovic2017graph}
\bibfield{author}{\bibinfo{person}{Petar Veli{\v{c}}kovi{\'c}},
  \bibinfo{person}{Guillem Cucurull}, \bibinfo{person}{Arantxa Casanova},
  \bibinfo{person}{Adriana Romero}, \bibinfo{person}{Pietro Lio}, {and}
  \bibinfo{person}{Yoshua Bengio}.} \bibinfo{year}{2018}\natexlab{}.
\newblock \showarticletitle{Graph attention networks}. In
  \bibinfo{booktitle}{\emph{ICLR}}.
\newblock


\bibitem[\protect\citeauthoryear{Wang, Cao, Wang, Sheng, Orgun, and Lian}{Wang
  et~al\mbox{.}}{2021}]%
        {wang2021survey}
\bibfield{author}{\bibinfo{person}{Shoujin Wang}, \bibinfo{person}{Longbing
  Cao}, \bibinfo{person}{Yan Wang}, \bibinfo{person}{Quan~Z Sheng},
  \bibinfo{person}{Mehmet~A Orgun}, {and} \bibinfo{person}{Defu Lian}.}
  \bibinfo{year}{2021}\natexlab{}.
\newblock \showarticletitle{A survey on session-based recommender systems}. In
  \bibinfo{booktitle}{\emph{ACM Computing Surveys}}, Vol.~\bibinfo{volume}{54}.
  \bibinfo{pages}{1--38}.
\newblock


\bibitem[\protect\citeauthoryear{Wang, He, Cao, Liu, and Chua}{Wang
  et~al\mbox{.}}{2019a}]%
        {wang2019kgat}
\bibfield{author}{\bibinfo{person}{Xiang Wang}, \bibinfo{person}{Xiangnan He},
  \bibinfo{person}{Yixin Cao}, \bibinfo{person}{Meng Liu}, {and}
  \bibinfo{person}{Tat-Seng Chua}.} \bibinfo{year}{2019}\natexlab{a}.
\newblock \showarticletitle{Kgat: Knowledge graph attention network for
  recommendation}. In \bibinfo{booktitle}{\emph{SIGKDD}}.
  \bibinfo{pages}{950--958}.
\newblock


\bibitem[\protect\citeauthoryear{Wang, He, Wang, Feng, and Chua}{Wang
  et~al\mbox{.}}{2019b}]%
        {wang2019neural}
\bibfield{author}{\bibinfo{person}{Xiang Wang}, \bibinfo{person}{Xiangnan He},
  \bibinfo{person}{Meng Wang}, \bibinfo{person}{Fuli Feng}, {and}
  \bibinfo{person}{Tat-Seng Chua}.} \bibinfo{year}{2019}\natexlab{b}.
\newblock \showarticletitle{Neural graph collaborative filtering}. In
  \bibinfo{booktitle}{\emph{SIGIR}}. \bibinfo{pages}{165--174}.
\newblock


\bibitem[\protect\citeauthoryear{Wang, Jin, Zhang, He, Xu, and Chua}{Wang
  et~al\mbox{.}}{2020}]%
        {wang2020disentangled}
\bibfield{author}{\bibinfo{person}{Xiang Wang}, \bibinfo{person}{Hongye Jin},
  \bibinfo{person}{An Zhang}, \bibinfo{person}{Xiangnan He},
  \bibinfo{person}{Tong Xu}, {and} \bibinfo{person}{Tat-Seng Chua}.}
  \bibinfo{year}{2020}\natexlab{}.
\newblock \showarticletitle{Disentangled graph collaborative filtering}. In
  \bibinfo{booktitle}{\emph{SIGIR}}. \bibinfo{pages}{1001--1010}.
\newblock


\bibitem[\protect\citeauthoryear{Xia, Wu, Wang, Chen, and Li}{Xia
  et~al\mbox{.}}{2022}]%
        {xia2022progcl}
\bibfield{author}{\bibinfo{person}{Jun Xia}, \bibinfo{person}{Lirong Wu},
  \bibinfo{person}{Ge Wang}, \bibinfo{person}{Jintao Chen}, {and}
  \bibinfo{person}{Stan~Z Li}.} \bibinfo{year}{2022}\natexlab{}.
\newblock \showarticletitle{ProGCL: Rethinking Hard Negative Mining in Graph
  Contrastive Learning}. In \bibinfo{booktitle}{\emph{ICML}}.
  \bibinfo{pages}{24332--24346}.
\newblock


\bibitem[\protect\citeauthoryear{Yang, Huang, Xia, and Li}{Yang
  et~al\mbox{.}}{2022}]%
        {yang2022knowledge}
\bibfield{author}{\bibinfo{person}{Yuhao Yang}, \bibinfo{person}{Chao Huang},
  \bibinfo{person}{Lianghao Xia}, {and} \bibinfo{person}{Chenliang Li}.}
  \bibinfo{year}{2022}\natexlab{}.
\newblock \showarticletitle{Knowledge Graph Contrastive Learning for
  Recommendation}. In \bibinfo{booktitle}{\emph{SIGIR}}.
  \bibinfo{pages}{1434--1443}.
\newblock


\bibitem[\protect\citeauthoryear{Zhang, Yao, Sun, and Tay}{Zhang
  et~al\mbox{.}}{2019}]%
        {zhang2019deep}
\bibfield{author}{\bibinfo{person}{Shuai Zhang}, \bibinfo{person}{Lina Yao},
  \bibinfo{person}{Aixin Sun}, {and} \bibinfo{person}{Yi Tay}.}
  \bibinfo{year}{2019}\natexlab{}.
\newblock \showarticletitle{Deep learning based recommender system: A survey
  and new perspectives}. In \bibinfo{booktitle}{\emph{ACM Computing Surveys}},
  Vol.~\bibinfo{volume}{52}. \bibinfo{pages}{1--38}.
\newblock


\bibitem[\protect\citeauthoryear{Zhuang, Liu, Zhang, Tan, Wu, Liu, Wei, Gu,
  Zhang, Zhou, et~al\mbox{.}}{Zhuang et~al\mbox{.}}{2020}]%
        {zhuang2020hubble}
\bibfield{author}{\bibinfo{person}{Chenyi Zhuang}, \bibinfo{person}{Ziqi Liu},
  \bibinfo{person}{Zhiqiang Zhang}, \bibinfo{person}{Yize Tan},
  \bibinfo{person}{Zhengwei Wu}, \bibinfo{person}{Zhining Liu},
  \bibinfo{person}{Jianping Wei}, \bibinfo{person}{Jinjie Gu},
  \bibinfo{person}{Guannan Zhang}, \bibinfo{person}{Jun Zhou}, {et~al\mbox{.}}}
  \bibinfo{year}{2020}\natexlab{}.
\newblock \showarticletitle{Hubble: An industrial system for audience expansion
  in mobile marketing}. In \bibinfo{booktitle}{\emph{SIGKDD}}.
  \bibinfo{pages}{2455--2463}.
\newblock


\bibitem[\protect\citeauthoryear{Zou, Wei, Mao, Wang, Qiu, Zhu, and Cao}{Zou
  et~al\mbox{.}}{2022}]%
        {zou2022multi}
\bibfield{author}{\bibinfo{person}{Ding Zou}, \bibinfo{person}{Wei Wei},
  \bibinfo{person}{Xian-Ling Mao}, \bibinfo{person}{Ziyang Wang},
  \bibinfo{person}{Minghui Qiu}, \bibinfo{person}{Feida Zhu}, {and}
  \bibinfo{person}{Xin Cao}.} \bibinfo{year}{2022}\natexlab{}.
\newblock \showarticletitle{Multi-level Cross-view Contrastive Learning for
  Knowledge-aware Recommender System}. In \bibinfo{booktitle}{\emph{SIGIR}}.
  \bibinfo{pages}{1358--1368}.
\newblock


\end{thebibliography}

%%
%% If your work has an appendix, this is the place to put it.

\end{document}